\pdfoutput=1

\documentclass[graybox]{svmult}
\usepackage{tcolorbox}
\usepackage{comment}
\usepackage{times}
\usepackage{latexsym}
\usepackage{soul}
\usepackage{url}
\usepackage[utf8]{inputenc}
\usepackage[font=small]{caption}
\usepackage{graphicx}
\usepackage{amsmath}
\usepackage{algorithm}
\usepackage{algorithmic}
\usepackage{booktabs}
\usepackage{multirow,siunitx,booktabs}
\usepackage{ulem}
\usepackage{paralist}
\usepackage{mathptmx}       
\usepackage{helvet}         
\usepackage{courier}        
\usepackage{type1cm}        
\usepackage{makeidx}         
\usepackage{graphicx}        
\usepackage[bottom]{footmisc}
\title*{ASMR: Augmenting Life Scenario using Large Generative Models for Robotic Action Reflection}
\author{Shang-Chi Tsai\quad Seiya Kawano\quad Angel Garcia Contreras\quad Koichiro Yoshino\quad Yun-Nung Chen}
\institute{Shang-Chi Tsai \at National Taiwan University, \email{d08922014@ntu.edu.tw}
\and Seiya Kawano \at RIKEN, \email{seiya.kawano@riken.jp}
\and Angel Garcia Contreras \at RIKEN, \email{angel.garciacontreras@riken.jp}
\and Koichiro Yoshino \at RIKEN, \email{koichiro.yoshino@riken.jp}
\and Yun-Nung Chen \at National Taiwan University, \email{y.v.chen@ieee.org}}
\begin{document}
\maketitle

\abstract{ 
When designing robots to assist in everyday human activities, it is crucial to enhance user requests with visual cues from their surroundings for improved intent understanding.
This process is defined as a multimodal classification task. 
However, gathering a large-scale dataset encompassing both visual and linguistic elements for model training is challenging and time-consuming.
To address this issue, our paper introduces a novel framework focusing on data augmentation in robotic assistance scenarios, encompassing both dialogues and related environmental imagery. This approach involves leveraging a sophisticated large language model to simulate potential conversations and environmental contexts, followed by the use of a stable diffusion model to create images depicting these environments.
The additionally generated data serves to refine the latest multimodal models, enabling them to more accurately determine appropriate actions in response to user interactions with the limited target data.
Our experimental results, based on a dataset collected from real-world scenarios, demonstrate that our methodology significantly enhances the robot's action selection capabilities, achieving the state-of-the-art performance.
}

\section{Introduction}
\label{sec:1}
As robotic technology advances, robots are increasingly capable of providing a variety of services in different contexts. A key challenge in robotics is understanding and responding to human requests that are not always clear-cut, especially in life-support situations~\cite{tanaka2021arta}. 
This requires interpreting both the visual context of the environment and the user's verbal communication, essentially making it a multimodal, multi-class classification problem. Recent advancements have seen the development of large multimodal language models~\cite{brohan2023rt2,driess2023palme,huang2023language,liu2023visual,peng2023kosmos2}, that can process and respond to multiple channels of input data.

Our study utilizes the latest multimodal model, LLaVA~\cite{liu2023visual}, as a foundation for predicting responsive actions to human requests. While LLaVA has shown promise in general multimodal interactions, it requires additional data to tailor its responses to specific actions in a human-robot interaction context. Gathering such interaction data is often time-intensive and not easily scalable.

Inspired by the success of the large generative model in the language~\cite{zhao2023survey} and vision~\cite{yang2023diffusion,zhang2023survey,zhang2023texttoimage} domains, this paper introduces a framework for automatically enhancing scenario data, specifically in contexts where a robot needs to perform life-support actions in response to human requests. 
We harness the power of large language models~\cite{openai2023gpt4,touvron2023llama,Bommasani2021FoundationModels,peng2023instruction}, to create plausible dialogue scenarios~\cite{kim-etal-2023-soda,huang2023converser,chang2023salesbot} and describe environmental settings. These narratives are then visualized using advanced diffusion models~\cite{li2023blipdiffusion,podell2023sdxl,rombach2022highresolution}, creating images that represent the robot's perspective during each dialogue.

By using this augmented scenario data, we can train an agent to choose appropriate actions based on everyday user interactions. This training is conducted in a controlled environment, supplemented with a small, real-world dataset collected from human-robot interactions. Our experiments demonstrate that this approach not only generates realistic scenarios but also effectively trains the multimodal language model to respond with appropriate life-support actions based on both verbal requests and environmental cues. The success of this framework highlights its potential to make robotic scenario data more scalable and relevant.

\section{Related Work}
\label{sec:2}
In this paper, we explore the augmentation of scenario data using large generative models. We provide an overview of the relevant background in large language models (LLMs) and stable diffusion models.

\subsection{Large Language Models (LLMs)}
Language models have been widely studied for research in language understanding and generation, evolving from statistical to neural network-based models~\cite{zhao2023survey}. In recent years, the emergence of pre-trained language models (PLMs)~\cite{devlin2019bert,raffel2023exploring,liu2019roberta,huang2022plmicdautomaticicdcoding} marked a significant advancement. These models, based on Transformer architecture and trained on vast text corpora, have shown remarkable proficiency across various natural language processing tasks. 
A key finding in this domain is that increasing model size enhances performance. As a result, the term ``large language models'' (LLMs) has been adopted to describe PLMs of substantial scale~\cite{floridi2020gpt}. A notable example is ChatGPT~\cite{ouyang2022training}, which has set new benchmarks in NLP tasks and demonstrates advanced linguistic capabilities in human interactions. The ongoing development and diversification of LLMs across various parameter sizes continue to be a focal point in both academic and industrial research~\cite{chowdhery2022palm,thoppilan2022lamda,touvron2023llama,chung2022scaling,tsai-chen-2025-balancing}.

\subsection{Large Diffusion Models}

Text-to-image generation has been a significant challenge in the field of computer vision~\cite{zhang2023texttoimage}.
Early attempts, such as AlignDRAW~\cite{mansimov2016generating}, produced images from text but lacked realism. 
The introduction of Text-conditional GANs~\cite{reed2016generative} marked a shift towards more sophisticated models capable of generating images from text descriptions, which is the first end-to-end architecture with characters as its input and pixels as its output.
However, these GAN-based methods were limited to smaller datasets. The advent of large-scale data utilization in autoregressive models, exemplified by DALL-E~\cite{ramesh2021zeroshot} and Parti~\cite{yu2022scaling}, brought improvements but at the cost of high computational demands and sequential error accumulation.

Recently, diffusion models have emerged as the new benchmark in text-to-image generation. These models can be broadly categorized based on their operational domain: pixel space or latent space. 
Pixel-level approaches, like GLIDE~\cite{nichol2022glide} and Imagen~\cite{saharia2022photorealistic}, generate images directly from high-dimensional data. On the other hand, latent space methods, such as stable diffusion~\cite{rombach2022highresolution} and DALL-E 2~\cite{ramesh2022hierarchical}, involve compressing images into a lower-dimensional space before applying the diffusion model. This innovation in model design has significantly enhanced the quality and efficiency of text-to-image generation.

\begin{figure*}[t!]
\centering
\includegraphics[width=\linewidth]{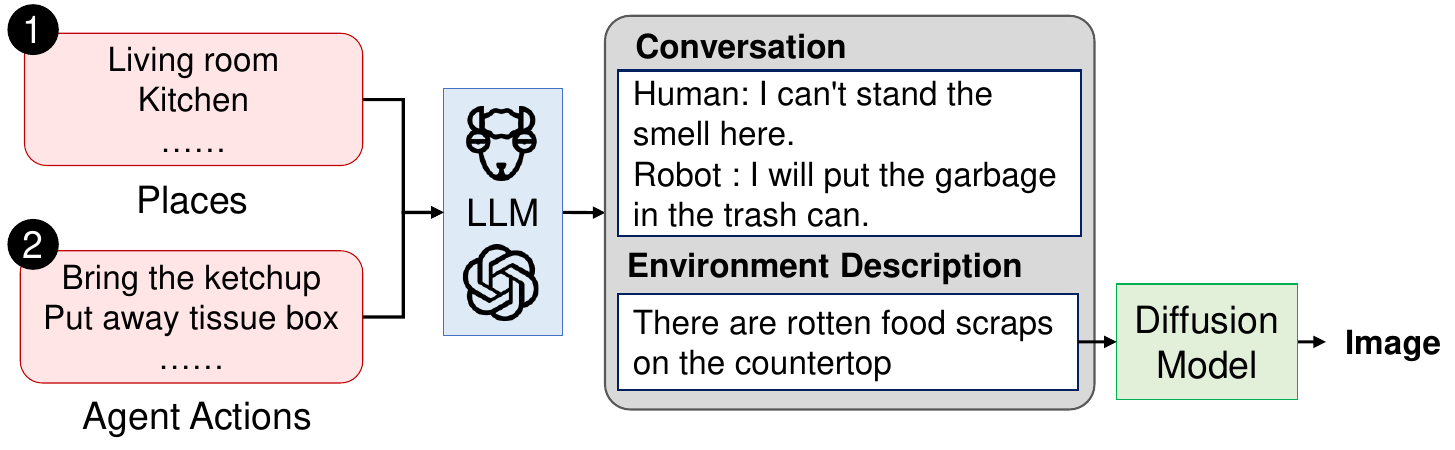}
\caption{Illustration of our augmentation method.}
\label{fig:augmentation}
\end{figure*}

\section{Proposed Augmentation Framework}
\label{sec:3}
In our framework, we approach the challenge of robotic action determination as a multi-class classification problem. The task involves interpreting an ambiguous request $\mathbf{x}$ from a user, coupled with an image depicting the robot's view of the environment. The objective is to accurately predict a suitable action $\mathbf{y} \subseteq \mathbf{Y}$, with $\mathbf{Y}$ representing the set of all actions available to the robot, to assist the human user effectively.

The primary challenge of training a model to tackle this task is the time-intensive and non-scalable nature of collecting authentic interaction data between humans and robots. To address this, we have developed a framework utilizing large generative models to enrich the dataset with various potential life-support scenarios, encompassing both dialogues and environmental images.
Figure~\ref{fig:augmentation} illustrates our augmentation pipeline.
Our framework comprises two distinct pathways, each tailored to generate robotic scenarios for a specific purpose.
\begin{itemize}
    \item \textbf{Place-based augmentation} focuses on creating dialogues pertinent to a specific location, such as a living room, kitchen, or bedroom, along with a detailed description of the respective environment.
    \item \textbf{Action-based augmentation} focuses on generating dialogues aligned with potential robot actions, like fetching a banana, clearing garbage, or organizing glasses, accompanied by a depiction of the setting where these actions would occur.
\end{itemize}

\begin{figure*}[t!]
\centering
\includegraphics[width=\linewidth]{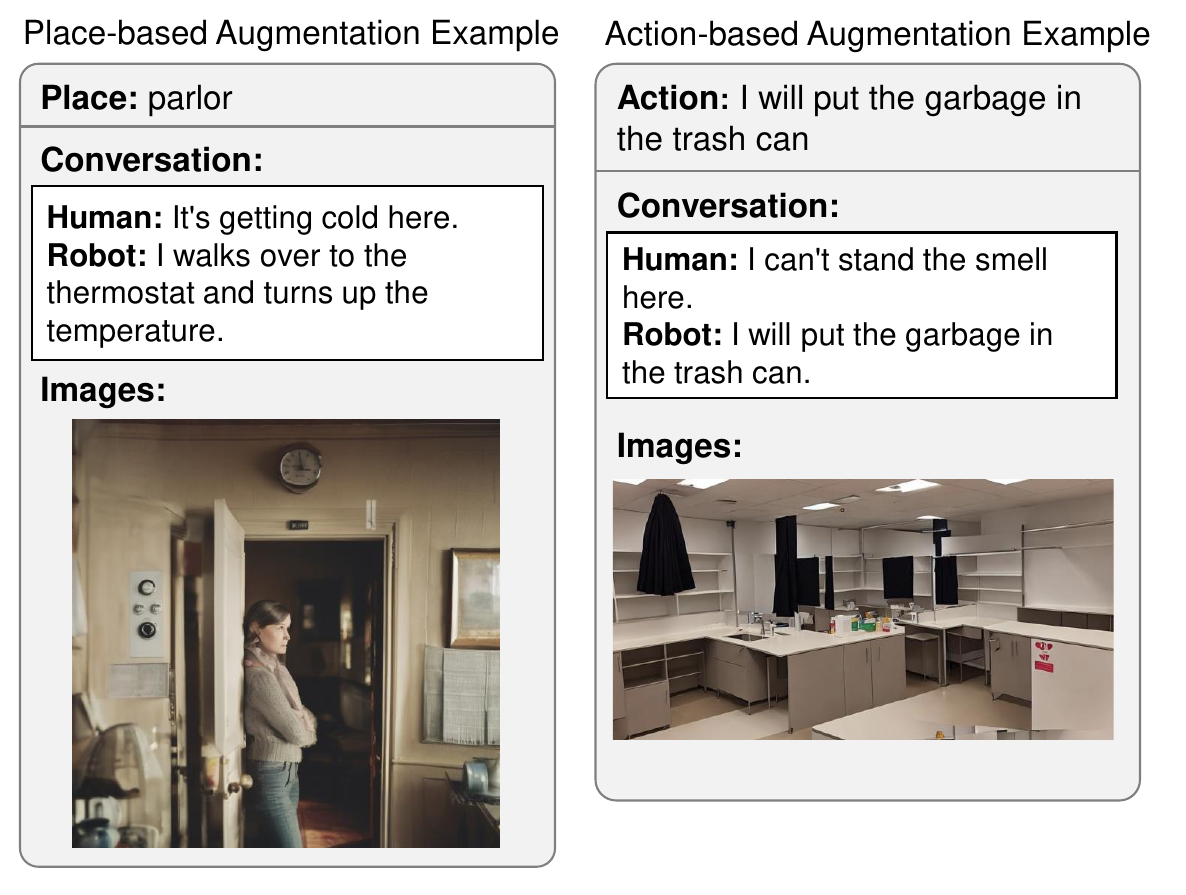}
\caption{Example of two augmentation methods.}
\vspace{-3mm}
\label{fig:example}
\end{figure*}

\subsection{Place-based Augmentation} 
\label{sec:placeAug}
In the initial phase of our augmentation pipeline, we employ gpt-3.5, a robust large language model, to create various dialogues. These dialogues simulate scenarios where a human presents an ambiguous request in everyday settings, and a robot must respond with an appropriate service action. The process begins by selecting a commonplace setting, such as a bedroom, bathroom, or dining room—areas where robots are likely to offer routine assistance. Next, we prompt gpt-3.5 to generate potential conversations that could occur in these settings, along with descriptions of the surrounding environment. Following this, we use the stable-diffusion-XL model~\cite{podell2023sdxl} to transform these textual descriptions into visual representations of the respective locations.

When crafting prompts for gpt-3.5, we do not set constraints on the generated user requests or robot actions. This approach allows the language model to conjure a wide array of scenarios, helping the model to learn and adapt to a diverse range of potential situations. For the image generation via the diffusion model, we emphasize the first-person perspective in the prompt, mirroring what the robot would observe in these environments.

We have identified ten everyday locations, each serving as the basis for generating ten distinct dialogues through the large language model. An illustrative example of this place-based augmentation process is depicted in the left part of Figure~\ref{fig:example}. The following is the prompt template used in our pipeline.

\begin{tcolorbox}[width=\columnwidth,colback=white]
\small
\begin{verbatim}
Give me ten conversation examples between two people in 
a [location]
Person A made an ambiguous request indirectly without asking
a question to Person B
And Person B responded with a reflected action to A
Each conversation should be one utterance
And describe some related object in the background 
\end{verbatim}
\end{tcolorbox}

\subsection{Action-based Augmentation}
\label{sec:actionAug}
While place-based augmentation focuses on equipping robots with the versatility to navigate various locations, action-based augmentation concentrates on creating scenarios tailored to specific, predefined robot actions~\cite{lai2022controllable}. 
In this second route of our framework, we utilize the same large language model as discussed in Section~\ref{sec:placeAug} for generating dialogues.

The key difference here lies in the nature of the input constraint. Rather than selecting a location, we choose an action from a robot's predefined action set, such as ``I will clean up the table''.
The gpt-3.5 model is then prompted to formulate potential dialogues where this action is the appropriate response, along with descriptions of the relevant surroundings. This approach allows the model to concentrate on learning and responding to specific, realistic scenarios tied to particular actions.

To generate images that resemble real-world settings, we employ the blip diffusion model~\cite{li2023blipdiffusion}, known for its ability to create images with a consistent theme or subject. When generating an image from a text description, we incorporate a reference image from our real-world data collection, specifying ``room'' as the constant subject. This method ensures the generated images closely align with the kind of environments a robot is likely to encounter.

Our framework includes 43 distinct actions, each serving as a basis to prompt the language model to produce ten unique dialogues. An example of this action-based augmentation is showcased in the right part of Figure~\ref{fig:example}. Below is a prompt template used with the large language model for this purpose.

\begin{tcolorbox}[width=\columnwidth,colback=white]
\small
\begin{verbatim}
Here is a reflected action from B. 
B: [reflected_action]
A is another person talking to B in a room
What ambiguous request may A talk to B indirectly without 
asking a question causing B to respond above reflected action.
And describe some related object in the background according 
to utterance between A and B 
\end{verbatim}
\end{tcolorbox}

After obtaining the scenario data derived from both the place-based and action-based augmentation routes, we construct our comprehensive augmented dataset.
This enriched dataset is then utilized to refine the performance of our base multimodal model, specifically designed for predicting robotic action responses. The model, adept at processing both visual and linguistic inputs, is trained to recognize and understand the scenarios presented in our augmented dataset.
Upon fine-tuning this base model, we proceed to assess its proficiency in zero-shot accuracy using real-world data. This evaluation helps us measure the model's ability to accurately predict robot actions in previously unseen situations, indicating the effectiveness of our augmentation approach.

\section{Experiments}
To assess the impact of our augmentation data, we conducted experiments using the Do-I-Demand dataset~\cite{10381769}, a collection of real interactive records between humans and robots. This dataset comprises 400 samples and serves as a benchmark for evaluating our method. We apply two base models with differing parameter sizes to test the efficacy of our proposed augmentation approach.

\subsection{Experimental Setup}
The evaluation dataset features two primary text elements: the human's ambiguous request and a description of the environment, inferred from an image. We develop two input settings based on these elements:
\begin{itemize}
    \item \textbf{Utterance}:
    Here, only the human's request is used as input, with the output being one of the 43 predefined actions.
    \item \textbf{Utter + Description}:
    This setting combines the human's request with the environmental description as input, aiming to predict one of the 43 predefined actions as output.
\end{itemize}

For our experiments, we select LLaVA, a large multimodal model renowned for its multimodal chat capabilities, as our base model. LLaVA integrates a vision encoder with a large language model (LLM) to facilitate general visual and linguistic understanding. We chose its two versions, 13B and 7B parameters, for subsequent fine-tuning.

We fine-tune the base models using our augmentation dataset for five epochs, keeping the hyperparameters largely consistent with those used in the original LLaVA model. The training input comprised the image and the ambiguous human request, with the goal of maximizing the likelihood of the model predicting the correct response action. This process was carried out on 4 A6000 GPUs, each with 40GB of memory, utilizing the LoRA technique~\cite{hu2021lora} for efficient training.

In evaluating the fine-tuned models, we focused on measuring zero-shot accuracy on the evaluation dataset. To match LLaVA's responses with specific actions, we employ a sentence encoder to process both the model's response and each potential action. We calculated the cosine similarity between each pair, selecting the action with the highest similarity as the final prediction. For this purpose, we experiment with two encoders: the Sentence-BERT model (SBERT)~\cite{reimers2019sentencebert} and the GPT-3 model~\cite{brown2020language}, both of which have shown excellent performance in various NLP benchmarks.

\subsection{Results}

The effectiveness of our augmentation methods on the Do-I-Demand dataset is summarized in Table~\ref{tab:main-result}. We evaluate the accuracy of each method by comparing the exact match rates across all labels.
The baseline results, achieved using the original multi-modal model LLaVA in two distinct sizes, are presented in the first row.
The data clearly indicates that both our place-based and action-based augmentation methods significantly enhance the performance of the base models. However, it is noteworthy that action-based augmentation generally outperforms place-based augmentation. This is likely because action-based augmentation is specifically tailored to align with the action categories in the evaluation dataset, whereas place-based augmentation aims to broadly improve the model's versatility in various scenarios.

Interestingly, we observe the highest performance when combining both place-based and action-based augmentations, except in one instance: the LLaVA-7B model with a GPT-3 encoder under the utterance-only setting. The top accuracy is recorded at 36.3\% for LLaVA-13B with the SBERT encoder in the utterance setting, and 48.8\% for LLaVA-7B with SBERT in the utterance plus description setting. These results reinforce the value of environmental descriptions in enhancing action prediction accuracy.


\begin{table}[t!]
\centering
\begin{tabular}{p{40mm}p{16mm}p{16mm}|p{16mm}p{16mm}}
\toprule
 \multirow{2}{*}{\bf Model} & \multicolumn{2}{c}{\bf Utterance-Only} & \multicolumn{2}{|c}{\bf Description + Utterance}  \\
 \cmidrule{2-5}
 & \hfil {\bf SBERT} & \hfil {\bf GPT3} & \hfil {\bf SBERT} & \hfil {\bf GPT3}  \\
\midrule
LLaVA-13B & \hfil 20.3 & \hfil 24.5  & \hfil 28.3 & \hfil 34.8  \\
~+ place-based augmentation & \hfil {29.0$^\dag$} & \hfil  {34.3$^\dag$}  & \hfil {33.3$^\dag$} & \hfil {39.0$^\dag$} \\
~+ action-based augmentation & \hfil {31.5$^\dag$} & \hfil {31.5$^\dag$}  & \hfil {45.5$^\dag$} & \hfil {45.5$^\dag$} \\
~+ both & \hfil {\bf 36.3$^\dag$} & \hfil {\bf35.3$^\dag$}  & \hfil {\bf 48.5$^\dag$} & \hfil {\bf 47.8$^\dag$} \\
\midrule
LLaVA-7B & \hfil 19.5 & \hfil 22.5  & \hfil 27.8 & \hfil 36.3 \\
~+ place-based augmentation & \hfil {30.3$^\dag$} & \hfil {\bf 36.1$^\dag$}  & \hfil {36.0$^\dag$} & \hfil {42.3$^\dag$}  \\
~+ action-based augmentation & \hfil {32.3$^\dag$} & \hfil {32.5$^\dag$}  & \hfil {41.8$^\dag$} & \hfil {41.5$^\dag$} \\ 
~+ both & \hfil {\bf 34.0$^\dag$} & \hfil {33.8$^\dag$} & \hfil {\bf 48.8$^\dag$} & \hfil {\bf 47.5$^\dag$} \\
\bottomrule
\end{tabular}
\caption{Results on the DO-I-DEMAND (\%). $^\dag$ indicates the significant improvement achieved by the augmented data. The best score for each base predictor is marked in bold.}
\label{tab:main-result} 
\end{table}

\subsection{Effectiveness with Diverse Prompts}
For place-based augmentation, our original prompt is ``make an ambiguous request indirectly without asking a question''. 
To explore variations, we test two alternative prompts: ``make an ambiguous request without asking a question'' and simply ``make an ambiguous request.''
After merging data generated from all three prompts, we observe a general decline in accuracy, as shown in Table~\ref{tab:third-result}. This suggests that our original prompt is sufficiently detailed, leading to the generation of high-quality dialogues for model training.


\begin{table}[t!]
\centering
\begin{tabular}{p{40mm}p{16mm}p{16mm}|p{16mm}p{16mm}}
\toprule
 \multirow{2}{*}{\bf Model} & \multicolumn{2}{c}{\bf Utterance-Only} & \multicolumn{2}{|c}{\bf Description + Utterance}  \\
 \cmidrule{2-5}
 & \hfil {\bf SBERT} & \hfil {\bf GPT3} & \hfil {\bf SBERT} & \hfil {\bf GPT3}  \\
\midrule
LLaVA-13B (original prompt) & \hfil 29.0 & \hfil 34.3  & \hfil 33.3 & \hfil 39.0 \\
~+ diverse prompts & \hfil 27.3 &  \hfil 32.0 & \hfil 35.0 & \hfil 43.5 \\
\midrule
LLaVA-7B (original prompt) & \hfil 30.3 & \hfil 36.1 & \hfil 36.0 & \hfil 42.3  \\
~+ diverse prompts & \hfil 23.3 & \hfil 25.5 & \hfil 34.0 & \hfil 42.0  \\
\bottomrule
\end{tabular}
\caption{Results of the original place-based and diverse prompts on the DO-I-DEMAND (\%).}
\label{tab:third-result} 
\end{table}

\subsection{Ablation of Blip Diffusion}
In our action-based augmentation, the use of the blip diffusion model for generating environmental images is crucial. We experiment by substituting blip diffusion with stable-diffusion-XL, as used in place-based augmentation. Table~\ref{tab:second-result} reveals a consistent decrease in accuracy across all scenarios, including a notable 6\% drop in the LLaVA-13B utterance setting. This highlights the significant role of the blip diffusion model in our augmentation strategy.


\begin{table*}[t!]
\centering
\begin{tabular}{p{40mm}p{16mm}p{16mm}|p{16mm}p{16mm}}
\toprule
 \multirow{2}{*}{\bf Model} & \multicolumn{2}{c}{\bf Utterance-Only} & \multicolumn{2}{|c}{\bf Description + Utterance}  \\
 \cmidrule{2-5}
 &\hfil {\bf SBERT} & \hfil {\bf GPT3} & \hfil {\bf SBERT} & \hfil {\bf GPT3}  \\
\midrule
LLaVA-13B (both augmentation) & \hfil 36.3 & \hfil 35.3  & \hfil 48.5 & \hfil 47.8 \\
~ w/o  blip diffusion & \hfil 30.5 & \hfil 31.8  & \hfil 46.0 & \hfil 45.5 \\
\midrule
LLaVA-7B (both augmentation) & \hfil 34.0 & \hfil 33.8 & \hfil 48.8 & \hfil 47.5 \\ 
~ w/o blip diffusion & \hfil 32.8 & \hfil 33.3 & \hfil 44.8 & \hfil 46.0 \\ 
\bottomrule
\end{tabular}
\caption{Results of the framework with and without blip diffusion on the DO-I-DEMAND (\%). }
\label{tab:second-result} 
\end{table*}

\subsection{Effectiveness on Low-Performing Labels}

Our analysis reveals that a significant portion, over one-third, of the actions predicted by the original multi-modal model perform zero accuracy. To delve into how our proposed augmentation methods impact these lower-performing labels, we group the labels into four categories based on their accuracy levels. 
Each group represents a quartile of performance, with bucket 1 consisting of the ten labels with the lowest accuracy, all at zero initially.

\begin{figure*}[t!]
\centering
\includegraphics[width=.8\linewidth]{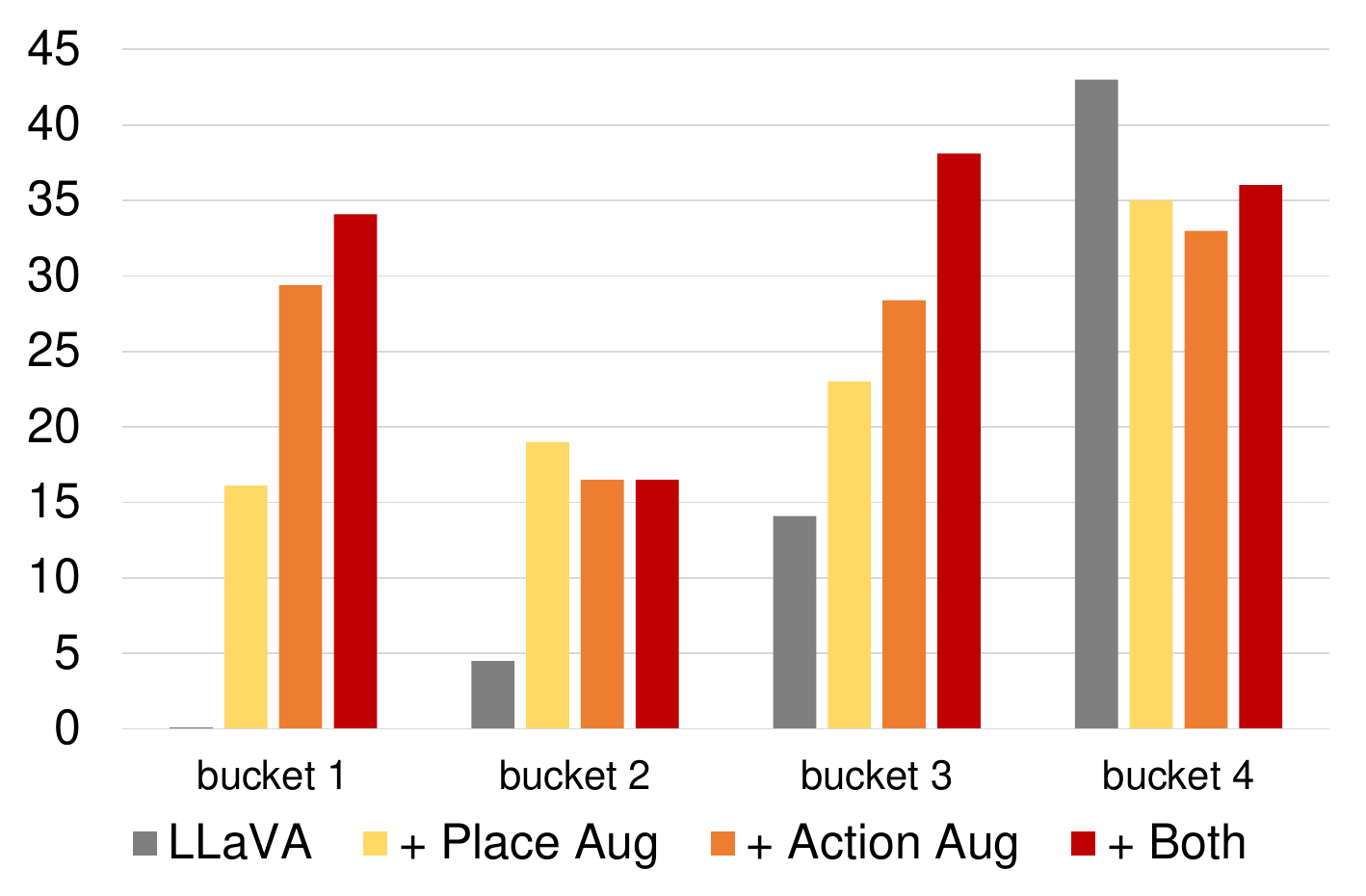}
\caption{performance of different buckets in utterance setting.}
\label{fig:bucket_utter}
\end{figure*}

In Figure~\ref{fig:bucket_utter}, we plot the mean performance of the LLaVA-1.5-13B model on the Do-I-Demand utterance set, categorized by label performance ranking. The graph clearly shows that our augmentation methods significantly improve the accuracy of labels in bucket 1. There is also a noticeable increase in accuracy across the other buckets.

\begin{figure*}[t!]
\centering
\includegraphics[width=.8\linewidth]{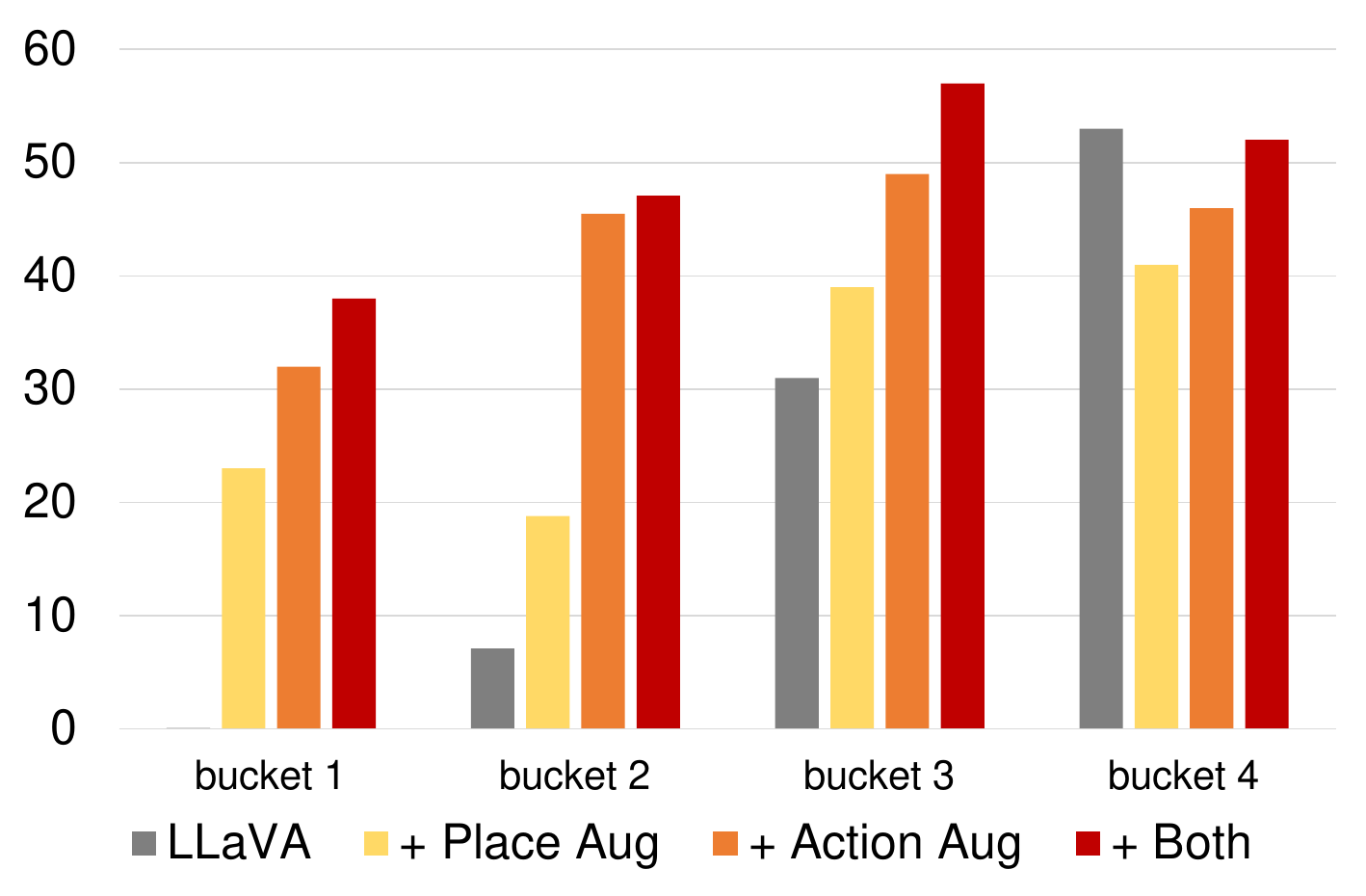}
\caption{performance of different buckets in utterance+description setting.}
\label{fig:bucket_utter_desc}
\end{figure*}

Similarly, Figure~\ref{fig:bucket_utter_desc} illustrates the mean performance on the Do-I-Demand utterance plus description set, again broken down by label performance ranking. This figure further confirms the positive impact of augmentation, especially in buckets 1 and 2, compared to the relatively lesser gains in buckets 3 and 4.

These findings underscore that each augmentation method we propose not only boosts overall performance but also effectively redistributes the performance across different labels, enhancing the model’s ability to predict a wide range of actions with improved accuracy.


\section{Conclusion}

In this paper, we introduce a novel pipeline designed to enhance the collection of robotic life-support scenario data, traditionally a time-consuming process. Our approach leverages a large language model to simulate dialogues between humans and robots, and a large diffusion model to create corresponding images of the environments.
We design two distinct types for dialogue generation: place-based augmentation, which focuses on scenarios occurring in specific places, and action-based augmentation, which centers around specific actions the robot might perform.
Both approaches have proven effective in generating realistic and relevant data, significantly aiding in the training of the LLaVA model. This model is fine-tuned to predict suitable actions based on ambiguous user requests and environmental imagery.
The experiments conducted on real-life collected data demonstrate that the augmented data not only significantly enhances the model's accuracy with all types of actions, especially low-performing ones, but also contributes to making robotic scenario data more scalable and adaptable. This advancement underscores the potential of our augmentation methods in enriching the training datasets for robotic action prediction models, thereby broadening their applicability in real-world scenarios.


%
%
%
\bibliographystyle{spmpsci}
\bibliography{custom}

\end{document}